\documentclass[10pt,twocolumn,letterpaper]{article}

\usepackage{iccv}
\usepackage{times}
\usepackage{epsfig}
\usepackage{graphicx}
\usepackage{amsmath}
\usepackage{amssymb}
\usepackage{subfig}

\usepackage{color}
\usepackage{soul}

\usepackage[breaklinks=true,bookmarks=false]{hyperref}

\iccvfinalcopy % *** Uncomment this line for the final submission

\ificcvfinal\fi

\begin{document}

%%%%%%%%% TITLE
\title{Appearance Codes using Joint Embedding Learning of Multiple Modalities}

\author{Evan Dogariu*\\
Princeton University\\
{\tt\small edogariu@princeton.edu}
\and
Alex Zhang*\\
Princeton University\\
{\tt\small alzhang@princeton.edu}
}

\maketitle
\ificcvfinal\thispagestyle{empty}\fi

%%%%%%%%% ABSTRACT
\begin{abstract}
    The use of appearance codes in recent work on generative modeling has enabled novel view renders with variable appearance and illumination, such as day-time and night-time renders of a scene. A major limitation of this technique is the need to re-train new appearance codes for every scene on inference, so in this work we address this problem proposing a framework that learns a joint embedding space for the appearance and structure of the scene by enforcing a contrastive loss constraint between different modalities. We apply our framework to a simple Variational Auto-Encoder model on the RADIATE dataset \cite{sheeny2021radiate} and qualitatively demonstrate that we can generate new renders of night-time photos using day-time appearance codes without additional optimization iterations. Additionally, we compare our model to a baseline VAE that uses the standard per-image appearance code technique and show that our approach achieves generations of similar quality without learning appearance codes for any unseen images on inference \footnote{Repository link: \hyperlink{https://github.com/edogariu/alex-zhang}{https://github.com/edogariu/alex-zhang}}.
\end{abstract}

%%%%%%%%% BODY TEXT
\section{Introduction}
In recent works on generative modeling and 3D neural rendering approaches, there has been an interest in dealing with dynamically changing scenes during the data collection process, especially because these techniques focus on generating static scenes. Additionally, there has been a new focus on developing interpretable forms of scene augmentation for generative models \cite{haque2023instructnerf2nerf, meng2022sdedit}. One solution proposed in \cite{bojanowski2019optimizing} is to jointly learn "appearance codes" for each scene that capture lighting and atmospheric appearance features during the rendering process. Both \cite{tancik2022blocknerf, DBLP:journals/corr/abs-2008-02268} use the same technique, and show that they are able to change the weather and time of day in their neural renderings using these appearance codes. However, a major flaw in this approach is the need to assign a new appearance embedding per render. Thus, on inference, these models require an extra optimization step on unseen renders to learn the relevant appearance embedding that can be interpolated for changing illumination conditions without affecting the 3D geometry of the scene. In our work, we aim to provide a different framework for learning appearance codes without the need for an extra inference optimization step.

% REST IS ADAPTED FROM PROPOSAL
The focus of the appearance code technique proposed in \cite{bojanowski2019optimizing} is to find a continuous low-dimensional latent space that can be interpolated in a way that does not affect the structural components of the scene. They explicitly enforce this constraint in \cite{tancik2022blocknerf, DBLP:journals/corr/abs-2008-02268} by only using the appearance codes as input to the branch of the NeRF that outputs color, and not the light density. To eliminate the need for extra optimization, a more interesting approach involves learning \textbf{joint embedding} given multiple input modalities, in which latent spaces are shaped in a specific way using certain modalities. For example, given access to LiDAR data and RGB data, structural information can be preserved in a latent space by applying contrastive learning over the output of the encoded LiDAR data (which provides depth information) and the encoded RGB data, while a separate latent space can be used for color \footnote{Joint embedding ideas were inspired by \hyperlink{https://atcold.github.io/NYU-DLSP21/en/week15/15-2/}{https://atcold.github.io/NYU-DLSP21/en/week15/15-2/}}.

We test our joint embedding approach idea using VAEs with several different encoders (all of which take the RGB data as input) that learn different information based on which modality they are being shaped by through contrastive learning. While VAEs tend to produce blurrier outputs, they are simpler to work with as a proof of concept to demonstrate that this idea can be generalized to larger frameworks in novel view synthesis and 3D generation. In this work, we make the following contributions:
\begin{enumerate}
    \item We propose a new joint-embedding framework that uses multiple modalities to enforce the learning of specific latent spaces for things like structure and appearance of the environment.
    \item We build a simple VAE architecture using our framework and demonstrate that it can generate day-time views of night-time images.
    \item We build a baseline VAE using the per-image appearance embedding approach proposed in \cite{bojanowski2019optimizing} and compare the outputs to our approach.
    \item We introduce a (to our knowledge) novel technique for decoupling latent embedding models that makes use of an anticontrastive loss. 
\end{enumerate}

\section{Related Works}
\subsection{Generative Models}
There are many techniques in the field of generative modeling and 2D/3D generation such as VAE \cite{kingma2022autoencoding}, GAN \cite{goodfellow2014generative, karras2019stylebased}, diffusion \cite{ho2020denoising}, etc., and many 3D methods that are designed based off of similar principles such as GET3D \cite{gao2022get3d} and GeNVS \cite{chan2023generative}. Each of these operates under a similar structure, where we apply deep learning to transform a known distribution such as a Gaussian to a generation distribution, potentially with the guidance and information of some priors. However, different techniques have different strengths and weaknesses. For example, the VAE generation procedure is known to learn latent representations that are fairly smooth, simple, and controllable/regularizable, while diffusion-based approaches are known to be able to generate with diversity and allow for excellent interpolation/editing in the latent space, though at the cost of inference speed. In our work we focus on VAEs for simplicity, but we note that other generative backbones can be used downstream as well.

\subsection{Appearance Codes} 
The general approach proposed in \cite{bojanowski2019optimizing} is to learn a low-dimensional appearance code or embedding per image, each of which occupies latent space that allows for interpolation between two different images. Because each appearance embedding is optimized per-image, during inference, new appearance codes are learned through an additional optimization step on half of the image \cite{tancik2022blocknerf, DBLP:journals/corr/abs-2008-02268}. To ensure that the appearance codes are explicitly for appearance, the NeRF-based approaches use a separate input channel for the appearance codes that only affect the output color of the network and jointly train the appearance codes with the rendering model. In our work, we avoid the need for extra optimization steps by learning a global appearance latent space.

\subsection{Contrastive Learning}
To enforce similarity in the latent representations learned through different modalities, we perform contrastive learning between the latent vectors of these different modalities, much like what is performed in CLIP \cite{radford2021learning} between text and image samples. Contrastive learning approaches like InfoNCE \cite{oord2019representation} and supervised contrastive learning \cite{DBLP:journals/corr/abs-2004-11362} use a loss function that penalizes latent representations from different modalities of different scenes being too similar, thus shaping the latent space to learn common attributes between the different modalities. In the case of 2D/3D generative modeling, using a contrastive learning objective between two different modalities enforces certain common information like the structure of the scene to be preserved. 

\section{Method}
\subsection{Setting}
Given access to multiple complementary input modalities, each of which may contain types of information, our goal is to learn latent embeddings that disentangle the different types of information found in each modality. This allows for free modification, interpolation, and customization in the different latent spaces without compromising the quality and information retainment of each embedding.

As a concrete and specific example, suppose that our two input modalities are RGB streams and LiDAR data (a setup that is commonly found on modern autonomous vehicles). We note that the RGB data contains information regarding both the appearance of the scene and also the 3D structure of the scene, though a usable 3D understanding of the environment is difficult to glean from RGB data alone. By contrast, LiDAR data only contains 3D structural information, with no knowledge about the appearance of the environment. Since the applications of such imaging systems tend to use appearance and structural information separately downstream, it is useful to be able to encode them separately in a disentangled way. 

We note that our approach will be applicable whenever complementary data modalities are provided that share some information but also have their own unique capabilities. As such, we will detail the method in the abstract setting with data modalities $\mathcal{A}$ and $\mathcal{B}$; $\mathcal{A}$ has its own unique type of information captured (denoted $\mathcal{A}_a$), and there is also a type of information about the input that both modalities capture ($\mathcal{A}_s$ and $\mathcal{B}_s$). In the abstract setting, our formal goal is therefore to disentangle embeddings of modality $\mathcal{A}$ into its different types. We wish to produce two embedding spaces:
\begin{itemize}
    \item a map $\mathcal{A} \rightarrow \mathcal{A}_a$ that captures no information of type $\mathcal{A}_s$ 
    \item a map $\mathcal{A} \rightarrow \mathcal{A}_s$ that captures no information of type $\mathcal{A}_a$
\end{itemize}
In the motivating example given above, $\mathcal{A}$ is an RGB stream and $\mathcal{B}$ is LiDAR data. $\mathcal{A}_a$ is appearance information (color and shading), while $\mathcal{A}_s$ and $\mathcal{B}_s$ denote 3D structural information (depth). 
% Given access to multiple complementary modalities including camera views during training, our approach is to disentangle parts of the view being rendered such as the structure of the landscape and the appearance/lighting by explicitly learning different latent embeddings that are optimized using the modalities. 
\subsection{Architecture}
\begin{figure}[t]
\begin{center}
   \includegraphics[width=0.9\linewidth]{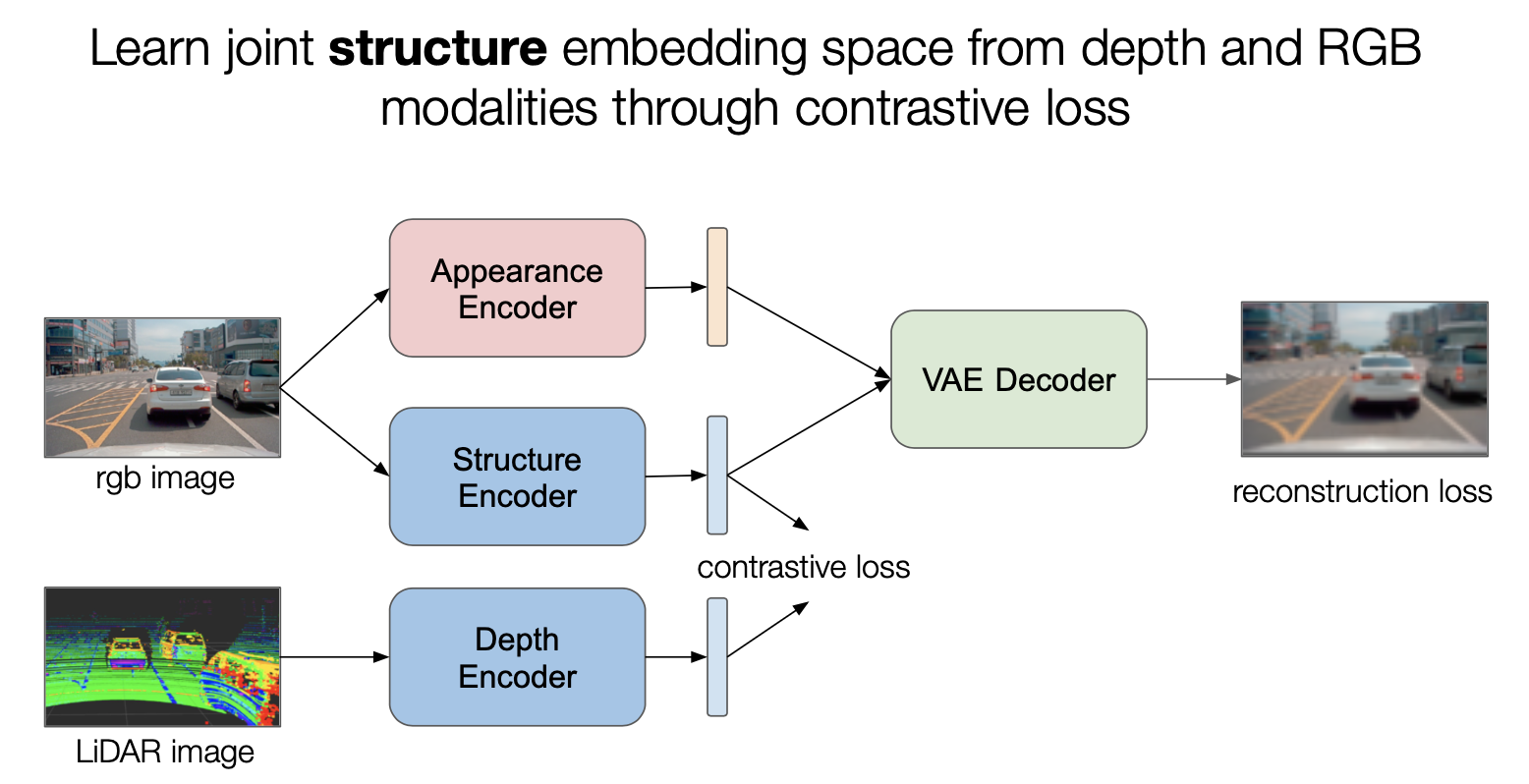}
\end{center}
   \caption{Our general approach is to learn separate latent spaces for the appearance and structure of the scene by relating different modalities through contrastive loss. For example, a joint embedding space of LiDAR depth information and RGB camera information can be learned using contrastive loss to explicitly consider depth-related information for RGB camera inputs. }
\label{fig:teaser}
\end{figure}

We use a VAE approach to learning latent spaces. In particular, we create three different encoders:
\begin{itemize}
    \item an encoder $E_a^{\mathcal{A}}: \mathcal{A} \rightarrow \mathcal{A}_a$
    \item an encoder $E_s^{\mathcal{A}}: \mathcal{A} \rightarrow \mathcal{A}_s$
    \item an encoder $E_s^{\mathcal{B}}: \mathcal{B} \rightarrow \mathcal{B}_s$
\end{itemize}
In order to enforce that the latent embeddings of modality $\mathcal{A}$ encode the required information, we also construct a decoder  $D^{\mathcal{A}}: \mathcal{A}_a \times \mathcal{A}_s \rightarrow \mathcal{A}$; in other words, the decoder takes latent embeddings of each information type from modality $\mathcal{A}$ as input and attempts to reconstruct the original data of modality $\mathcal{A}$. A schematic of this setup is provided in Figure \ref{fig:teaser}

\subsection{Training Procedure}
Let $i$ denote indices of data points within the dataset, and let $B$ denote a set of indices contained within a batch.
For a single multimodal data point $\left(x^{\mathcal{A}}_i, x^{\mathcal{B}}_i\right) \in \mathcal{A} \times \mathcal{B}$, we apply our encoders to obtain three embeddings 
$$z_{a; i}^{\mathcal{A}} := E_a^{\mathcal{A}}(x_i^{\mathcal{A}}), \; z_{s; i}^{\mathcal{A}} := E_s^{\mathcal{A}}(x_i^{\mathcal{A}}), \; z_{s; i}^{\mathcal{B}} := E_s^{\mathcal{B}}(x_i^{\mathcal{B}})$$
We also apply our decoder to obtain a reconstruction $$\widehat{x}_{a; i}^{\mathcal{A}} := D^{\mathcal{A}}\left(z_{a; i}^{\mathcal{A}} \oplus z_{s; i}^{\mathcal{A}}\right),$$
where $\oplus$ denotes concatenation.

% To enforce disentanglement of the latent representations, we apply (1) a contrastive loss to $(z_s^{\mathcal{A}}, z_s^{\mathcal{B}})$ to ensure that they look the same and (2) an anticontrastive loss to ensure that $(z_a^{\mathcal{A}}, z_s^{\mathcal{A}})$ are as uncorrelated as they can be. In addition, we apply a reconstruction loss between $x^{\mathcal{A}}$ and $D^{\mathcal{A}}\left(z_a^{\mathcal{A}} \oplus z_s^{\mathcal{A}}\right)$ to enforce usefulness of the latent representations.
% \newline \newline

To enforce that $E^{\mathcal{A}}_s$ and $E^{\mathcal{B}}_s$ learn a joint understanding of the shared information type, we follow \cite{DBLP:journals/corr/abs-2004-11362} to compute an efficient batched contrastive loss objective: for a batch $B$ of indices, we have
$$ \mathcal{L}_{con} :=  - \sum_{i \in B} \log \frac{\exp \left(z_{s; i}^{\mathcal{A}} \cdot z_{s; i}^{\mathcal{B}} / \tau\right)}{\sum_{j \in B \setminus \{i\}} \exp\left(z_{s; i}^{\mathcal{A}} \cdot z_{s; j}^{\mathcal{B}} / \tau\right)},$$
where $\tau$ is a temperature parameter.
This prefers encoders $E^{\mathcal{A}}_s$ and $E^{\mathcal{B}}_s$ that encode matched pairs $\left(x^{\mathcal{A}}_i, x^{\mathcal{B}}_i\right) $ to similar embeddings of the shared type $s$, while pusing the encodings for unmatched pairs within the batch away.

To enforce that $E^{\mathcal{A}}_a$ and $E^{\mathcal{A}}_s$ remain as uncorrelated as possible (i.e. changing the information of type $\mathcal{A}_s$ shouldn't change the output of $E^{\mathcal{A}}_a$), we apply a (to our knowledge) novel technique for anticontrastive loss that prefers embeddings with low alignment. To be precise, for a batch $B$ of indices, we compute
$$\mathcal{L}_{anti}:= \frac{1}{2|B|}\sum_{i \in B} \left(\frac{z_{a; i}^{\mathcal{A}} \cdot z_{s; i}^{\mathcal{A}}}{||z_{a; i}^{\mathcal{A}}||_2 ||z_{s; i}^{\mathcal{A}}||_2}\right)^2$$
In other words, the anticontrastive loss is the mean squared cosine of the angle between the two embeddings of the modality $\mathcal{A}$, which is a measurement of either positive or negative alignment. Minimization of this loss in tandem with the other loss functions biases the $E^{\mathcal{A}}_a$ and $E^{\mathcal{A}}_s$ encoders to make use of disentangled types of information, which is exactly the desired behavior.

We additionally compute a perceptual reconstruction loss for the decoder output, which for a VAE is the MSE loss: for a batch of indices $B$,
$$ \mathcal{L}_{rec} = \frac{1}{2|B|} \sum_{i\in B} (x_i^{\mathcal{A}} - \hat{x}_i^{\mathcal{A}})^2  $$

Additionally, we use KL-divergence regularization on the VAE encoders for stability, defined on a batch of indices $B$ to be
$$\mathcal{L}_{KL} := -\frac{1}{2|B| \cdot d} \sum_{i \in B} \sum_{n = 1}^d \left(\mu_{i, j}^2 + \sigma_{i, j}^2 - \log \sigma_{i, j}^2 - 1\right),$$
where $\mu_{i, j}$ is the $j^{th}$ component of predicted mean vector, $\sigma_{i, j}$ is the predicted standard deviation, $d$ is the latent dimension. This formula is a simplified form of the negation of the KL divergence between the predicted distribution and the latent prior distribution (a unit normal), and is used to stabilize encoder training. We sum the contributions of this loss from all 3 encoders.

In total, our models are jointly learned to try to minimize
\begin{align*} \mathcal{L} = \mathcal{L}_{rec} &+ \lambda_{con} \mathcal{L}_{con} + \lambda_{anti} \mathcal{L}_{anti} + \lambda_{KL} \mathcal{L}_{KL} \end{align*}
for weighting hyperparameters $\lambda_{con}, \lambda_{anti},$ and $\lambda_{KL}$.

\section{Experiments}
\subsection{Dataset}
We run our experiments on two datasets of different scales, difficulties, and applicabilities. 

The first is the RGB-D Object Dataset \cite{rgbd}, which is composed of matched RGB and depth images of common tabletop objects and furniture. There are 14 distinct scenes, each with many RGB-D images from different views, totaling 11,427 datapoints. The scenes are fairly consistent (same lighting, same background room), and will offer an initial playground for our approach and a study of the effects of our joint embedding spaces.

The second, more challenging and more applicable dataset is the RADIATE \cite{sheeny2021radiate} dataset, which contains RGB camera and LiDAR scene captures of roads under various adverse weather (fog, snow, rain) and lighting conditions. We specifically focus on generating day-time versions of night-time scenes, but we also experiment with applying this technique to other adverse weather conditions. For our multiple modalities, we focus on depth information from the LiDAR sensors\footnote{We reproject the LiDAR point clouds to sparse 2D depth images from the same view angle as the RGB image taken at the nearest timestep and then apply grid interpolation to make them dense, but we emphasize that our approach will work for 3D LiDAR data as well (for a suitably chosen encoder architecture for $E^{\mathcal{B}}_s$).} and the RGB information from the cameras.  This dataset consists of about 3 hours of drive footage and multimodal sensor data in many different conditions, though for computational reasons we had to restrict the dataset size to 16,000 frames split equally between daytime and nightime.

\subsection{Models}
The three encoders are convolutional, each using 5 layers of batch-normed convolutional blocks with channel dimensions increasing from 32 multiplicatively to 512; the resulting feature representation is then flattened and linearly projected to a latent dimension of 128. A transposed convolutional decoder is used in which the concatenated appearance and structure embeddings are linearly projected to the convolutional output dimension and then used to generate images of the input size. More implementation information can be found in the attached code, which is provided \footnote{See \hyperlink{https://github.com/edogariu/alex-zhang}{https://github.com/edogariu/alex-zhang}}. The data is rescaled to input dimension $(128, 128)$ before being passed into the models.

\subsection{Hyperparameters}
We set the latent dimension sizes to be 128 for all encoders.
 We set a batch size of 64, and an initial learning rate of 0.0012 that decays by a factor of 0.8 every 50 epochs. We use the Adam optimizer with weight decay $1E-5$. We train for 400 epochs or until convergence, which is observed using a patience algorithm with patience 5 evaluated on validation data with respect to the reconstruction loss only. For loss weighting hyperparameters, we set $\lambda_{con} = 0.02, \lambda_{anti}=0.0005$, and $\lambda_{KL} = 5E-5$ when the respective losses are enabled. The above hyperparameters were all selected on a validation set, and are shared when training on both datasets. Also, we use a temperature parameter $\tau = 0.1$ for the contrastive loss, as is recommended in \cite{DBLP:journals/corr/abs-2004-11362}.

 \section{Results}
 \label{section:results}
\begin{figure}[t]
\begin{center}
   \includegraphics[width=0.9\linewidth]{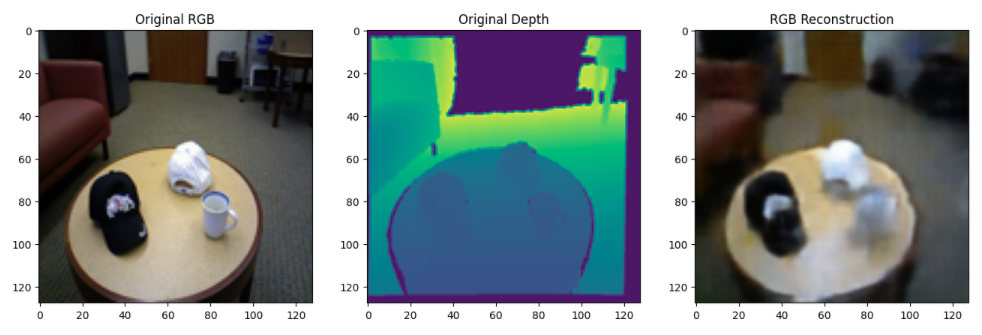}
\end{center}
   \caption{Example VAE reconstruction of a typical datapoint in the RGB-D dataset. We see that the reconstructed output is blurry, but is close enough that the encoders certainly learned useful latent representations.}
\label{fig:vanilla_rgbd_output}
\end{figure}

Before investigating the structure and topology of our latent spaces, it is important to ensure that the learned embeddings are useful. As seen in Figure \ref{fig:vanilla_rgbd_output}, while the outputs of the VAE are blurry, it is clear that there is enough information in the latent embeddings to reconstruct the image with, and so analysis of the latent spaces is useful. Reconstructions of images from the RADIATE dataset are of a very similar quality. 

Our main tools with which to analyze the latent spaces will be investigation into clustered embeddings, as well as experiments involving modification in the latent space. In Section \ref{section:ablation}, we compare these results against similar results when we disable the contrastive and anticontrastive losses to understand the effect of our approach.

\subsection{Clustering}
\begin{figure}[h]
\begin{center}
\subfloat[\centering Appearance Embeddings for RGB-D]{\includegraphics[width=0.45\linewidth]{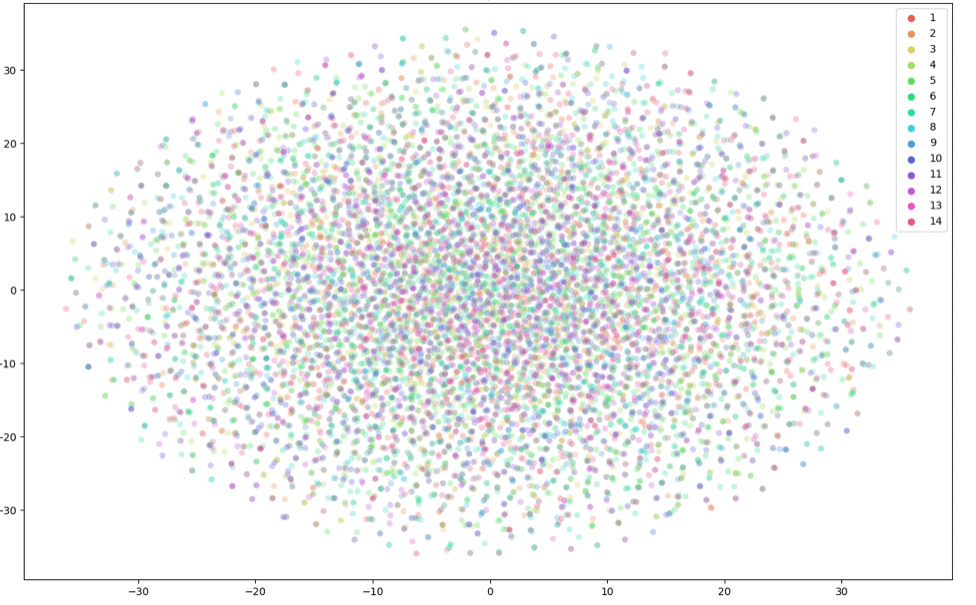}}
\subfloat[\centering Appearance Embeddings for RADIATE]{\includegraphics[width=0.45\linewidth]{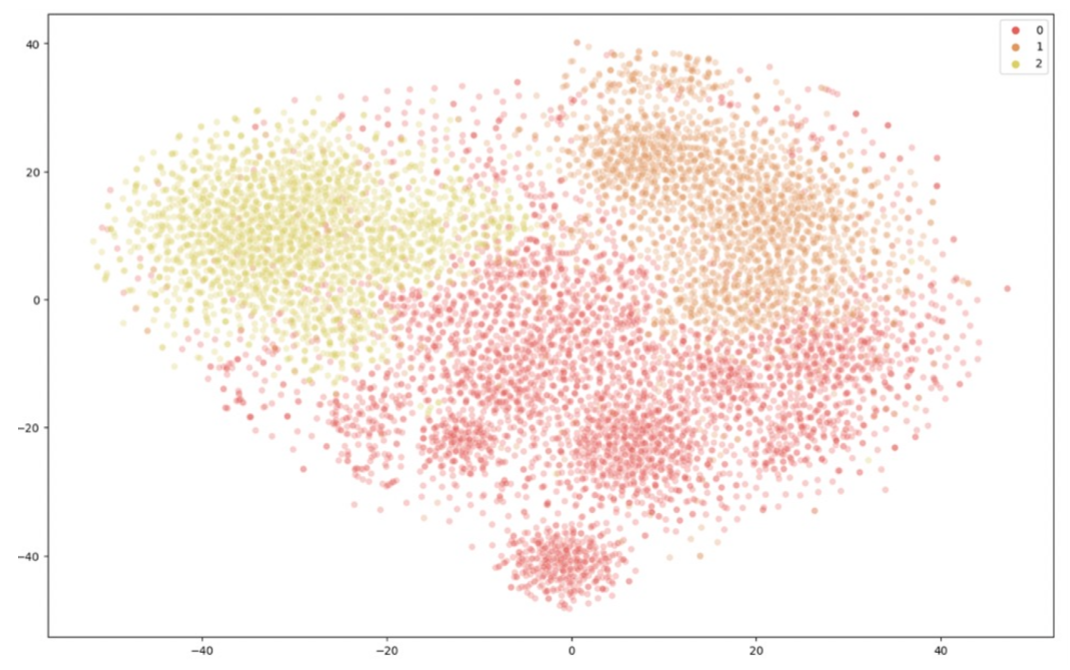}}
   
\end{center}
   \caption{t-SNE plots for \textbf{appearance} embedding spaces on both datasets. Color indicates scene label, which for RGB-D denotes individual scenes and for RADIATE denotes one of \{day-time, night-time, foggy\}.}
\label{fig:contrastive_appearance}
\end{figure}

In Figure \ref{fig:contrastive_appearance}, we visualize the appearance embedding spaces of our trained framework on both the RGB-D and RADIATE datasets\footnote{We make use of the t-SNE (t-distributed stochastic neighbor embedding) method to generate 2D visualizations of our latent space. This method produces visualizations that are similar to those produced by plotting the top 2 principal directions, but with more emphasis on ensuring the relative invariance of distance between points, making it an often better choice for visualizing clustering at the expense of distorted angles.}. On the RGB-D dataset, scenes look very similar with regard to appearance, as objects tend to have similar colors and the background is always colored and lit similarly. This is reflected in the fact that the appearance embeddings do not seem to cluster at all, and instead vary from data point to data point. On the RADIATE dataset, however, it is the case that since scenes from different lighting and weather conditions have distinct appearances, the embeddings cluster accordingly. When inferencing the model, it can be shown that for two RADIATE data points of a similar setting, one taken at day-time and one taken at night-time, the resulting embeddings are quite distinct \footnote{We leave testing of this sort to the reader for space considerations.}.

\begin{figure}[h]
\begin{center}
\subfloat[\centering Structure Embeddings for RGB-D]{\includegraphics[width=0.5\linewidth]{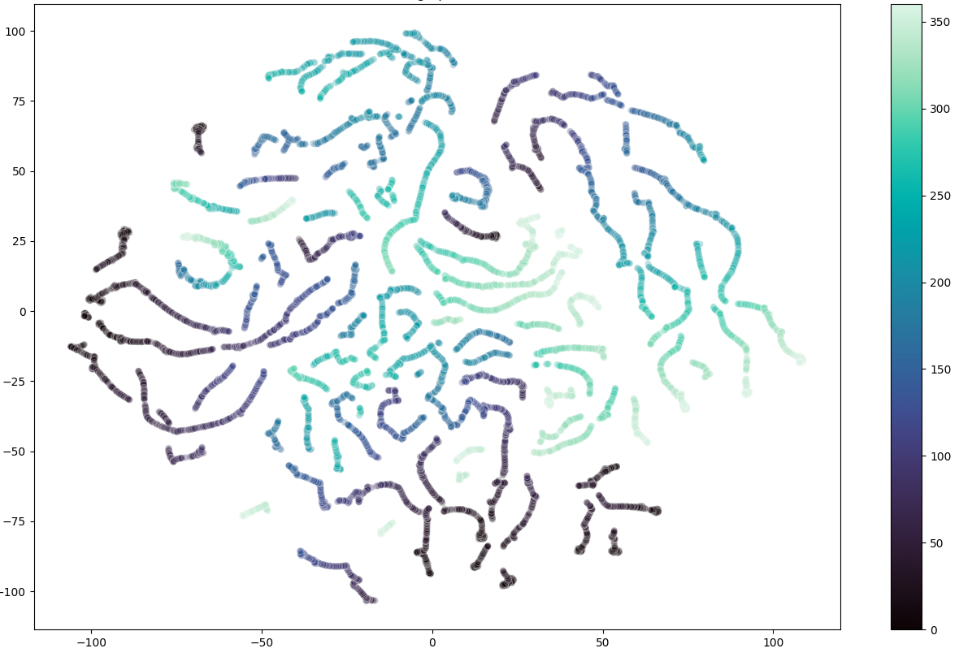}}
\subfloat[\centering Structure Embeddings for RADIATE]{\includegraphics[width=0.45\linewidth]{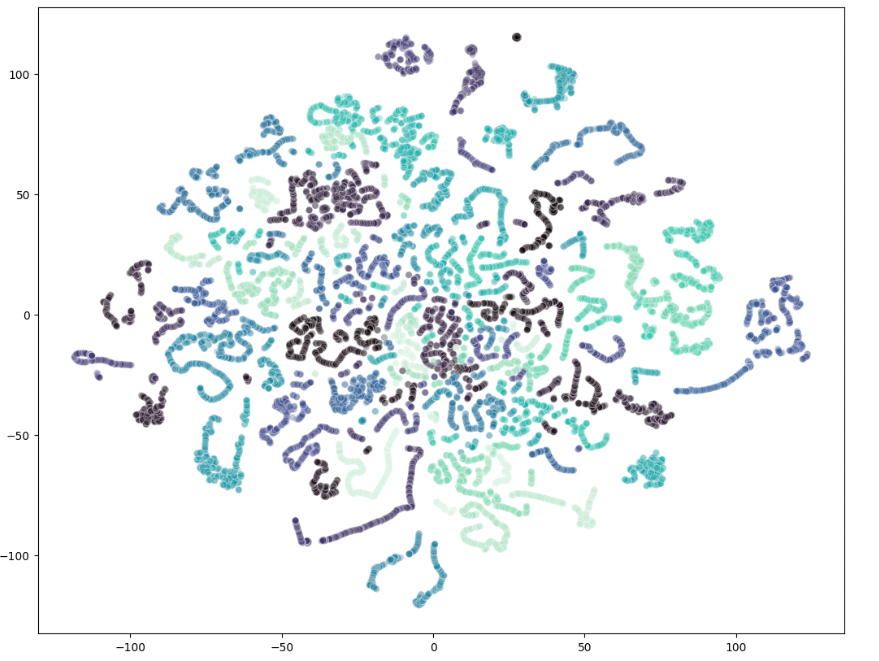}}
   
\end{center}
   \caption{t-SNE plots for \textbf{structure} embedding spaces on both datasets. Color indicates view direction for RGB-D, while for RADIATE it indicates location that the drive took place.}
\label{fig:contrastive_structure}
\end{figure}

Next, we visualize the structure latent space learned by our approach\footnote{Due to the application of our contrastive loss, the visualizations look identical for $z_{s}^{\mathcal{A}}$ and $ z_{s}^{\mathcal{B}}$ for both datasets (up to a rotation), and so we only display $z_{s}^{\mathcal{A}}$ to conserve space.}. In Figure \ref{fig:contrastive_structure}, we find that the latent space is of a qualitatively different shape to that of the appearance embeddings, with much finer clusters appearing. Upon further investigation, the clusters that form in the RGB-D dataset correlate very heavily with the view direction of the image, as seen by the continuity of the clusters under continuous change of the view direction. In the RADIATE dataset, clusters form instead according to the map location of each data point, with different clusters corresponding to different drives of the test vehicle. These are examples of \textbf{structural} information, which seems to highly inform the topology of the structure latent spaces without having significant effect on the topology of the appearance latent spaces (compare with what is seen in Figure \ref{fig:contrastive_appearance}). 

\subsection{Modification in the Latent Space}
\begin{figure*}
\begin{center}
\includegraphics[width=0.9\linewidth]{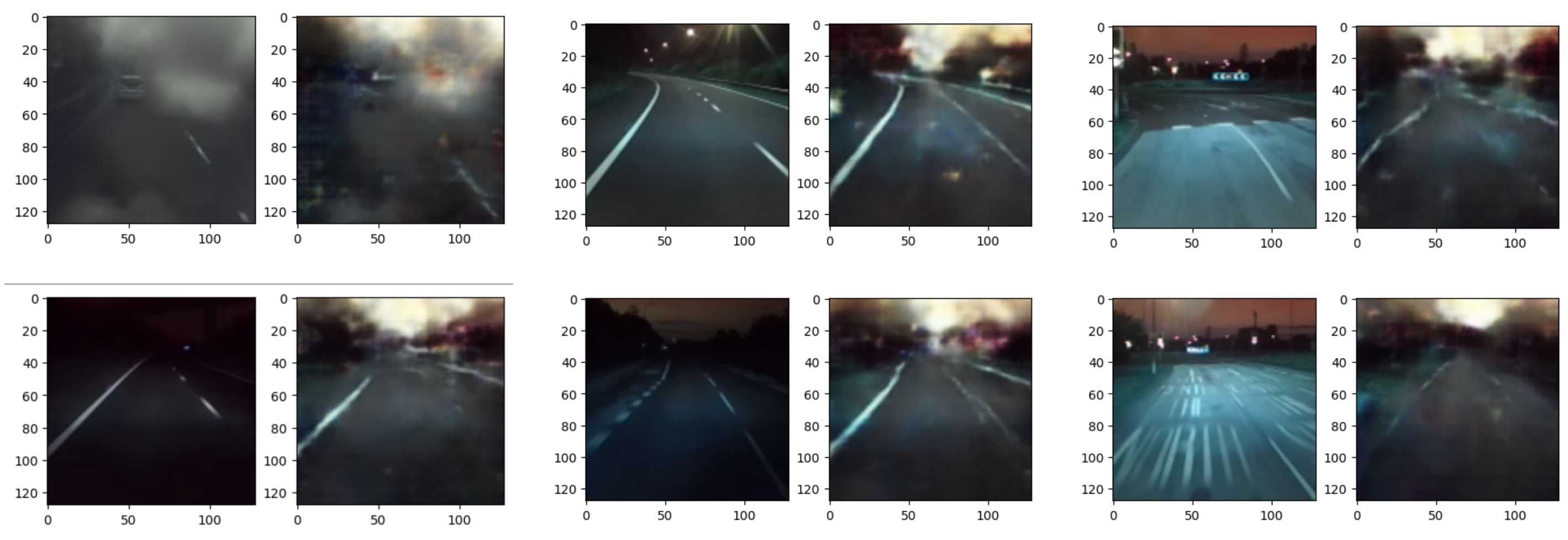}
\end{center}
   \caption{Results of our framework on a VAE doing inference on night-time images to generate day-time versions (left is original, right is generated). We use a "mean" of all appearance encoder outputs on day-time images in the train-set as the appearance encoding on inference. Observe that for out-of-distribution images like the right-most pairs, the generations contain many artifacts. For fog (top left), a similar effect is seen, but the day-time appearances are clear.}
\label{fig:results}
\end{figure*}
In this section, we attempt to make use of our disentangled latent spaces to change one aspect of data points without affecting the other. In particular, we use RADIATE data points that were captured during the night-time and apply the mean day-time appearance embedding during the decoding process; the results of this experiment are shown in Figure \ref{fig:results}. The sky is seen to change in such a way that matches the day-time appearances (most day-time images were actually taken in late afternoon/early evening), but structural information (such as the direction of the road, the existence of fog, and the presence of a car in the top left generation) are retained. 

\section{Ablations \& Analysis}
\label{section:ablation}
\subsection{Baseline VAE}
We tried two different baseline VAEs: one that, for every image, learns a lookup embedding for the appearance and an embedding for the structure, and one that uses our model's encoder structure for the structure embedding and learns a lookup embedding for just the appearance embedding. Both VAEs are able to perform standard image reconstruction; however, we note that when using the inference optimization technique described by \cite{bojanowski2019optimizing}, the image outputs from interpolating between embeddings are not clear, as unlike in \cite{DBLP:journals/corr/abs-2008-02268}, we do not have similar scenes with different illumination conditions. We know by design that our approach can learn a mapping from any input to appearance embedding, and the fact that optimizing appearance embeddings directly does not work in our setting and indeed requires inputs of the same location under different appearances in the dataset. As such, our approach seems to be more widely applicable, since the learned mapping from input to embedding does not require multiple data points of the same input location.

\subsection{Removing Contrastive Losses}
In this section, we ask whether similar disentanglement of the appearance and structure appears without the contrastive and anticontrastive losses and simply by virtue of the fact that we have two different encoders for RGB data. Our findings suggest that the answer is no, and that the benefits we see in Section \ref{section:results} are a result of contrastive learning applied to our architecture. While the reconstructions are of a similar quality, we find that the appearance and structure latent spaces for RGB data look nearly identical, and make maximal use of both appearance and structure information. While this may be useful for forming the most compressed representations possible, there is still large entanglement between the different information types.

\section{Discussion}
Our results applying our framework on a VAE model suggest that the appearance and structure latent spaces learned by the respective encoders are disentangled quite clearly. This framework is useful in that the multiple modalities are only used to shape the different latent spaces; during inference, only modality $\mathcal{A}$ (the raw RGB camera input in our experiments) is needed, as the different encoders learn to focus on different information based on the modality contrastive learning is performed with. Furthermore, while our results are pretty low-resolution and blurry due to the nature of VAE training, we note that our use of contrastive learning to shape these joint embedding spaces is extremely general and would work well with other frameworks like GANs, diffusion models, or NeRFs. In addition, we emphasize that the goal of our approach is to \textbf{disentangle} the representations in order to make disentangled use of them downstream; as such, downstream uses of these disentangled embeddings will likely address tasks other than generation.

Unlike the standard appearance code technique proposed in \cite{bojanowski2019optimizing}, our technique does not require the model to have a well-optimized latent space for the appearance code of a certain scene (i.e. having seen a similar scene before) to transfer "day-time" appearances to a scene. Instead, we can interpolate between known day-time scenes in our training set and form appearance embeddings that can be used during inference without extra optimization to modify the rendered scene, as demonstrated in Figure \ref{fig:results}. More generally, these learned appearance embeddings can be learned at train time and used in other frameworks; this also alleviates the necessity for "appearance matching" at deployment time in works such as \cite{tancik2022blocknerf}. However, it is unclear whether a single "day-time" appearance embedding suffices across all outputs, as many of the scenes in the RADIATE dataset are similar, which may not be the case for other datasets. More complex interpolation in the latent space may be required; our approach, though, will almost certainly prove helpful. An important drawback of our approach is the computational and memory overhead of learning these extra models, which are small in comparison with large-scale models like \cite{tancik2022blocknerf, gao2022get3d} but are still non-negligible.

Finally, we only focus on two modalities (the standard RGB camera and depth information), but we note that this framework is extendable to an arbitrary number of different modalities. Having access to different modalities can be advantageous because under adverse conditions, certain modalities produce clearer outputs than others. Put differently, different modalities have different strenghts and weaknesses, and separating the embeddings in terms of their strengths and weaknesses can help design better systems. For example, in trying to de-noise a scene and produce clear output images for applications like self-driving vehicles, our learned joint embedding framework is extremely useful because we know that depth information may be more robust to certain types of adverse weather. For future work, it would be of interest to extend this to novel view synthesis problems or more advanced generation problems to see what kind of clear renderings we can produce, or perhaps to other downstream tasks for which disentangled representations prove useful or even necessary.

One limitation that is present in our approach and clearly seen in Figure \ref{fig:results} is that, when using the mean appearance embedding to do latent space modification, the mean appearance information appears to sort of override some of the structural information (style of road markings, treeline structure, etc.). A smarter method of latent space modification is likely needed, but we are confident that our approach will learn embedding spaces that allow for more to be done with them.

In conclusion, we have presented a methodology that, when applied to complementary data modalities that capture multiple types of information, allows for the disentanglement of the multiple types of information into distinct latent spaces for downstream use. Such a framework can be used to replace current technique of directly optimizing appearance codes, as it is applicable with less restrictions on datasets and more flexibility on inference. However, it comes at a cost of computation and memory which we hope will be small in the scope of its applcations and perhaps can be alleviated with further research. 

\section{Acknowledgements}
We thank Professor Felix Heide and Fangyin Wei for their support and organization of an amazing semester! Thank you! We also thank the makers of the RGB-D Objects and RADIATE datasets for kindly sharing their data with us.
 
{\small
\bibliographystyle{ieee_fullname}
\bibliography{egbib}
}

\end{document}